\documentclass{article}

\usepackage{microtype}
\usepackage{graphicx}
\usepackage{subcaption}
\usepackage{booktabs} 
\usepackage{makecell} 
\usepackage{hyperref}

\usepackage[accepted]{icml2025}

\usepackage{amsmath}
\usepackage{amssymb}
\usepackage{mathtools}
\usepackage{amsthm}
\usepackage[capitalize,noabbrev]{cleveref}
\DeclareMathOperator*{\ub}{ub}
\theoremstyle{plain}
\newtheorem{theorem}{Theorem}[section]

\theoremstyle{definition}
\newtheorem{definition}[theorem]{Definition}
\newtheorem{assumption}[theorem]{Assumption}
\theoremstyle{remark}

\usepackage[textsize=tiny]{todonotes}

\icmltitlerunning{Error Distribution Smoothing}

\begin{document}

\twocolumn[
\icmltitle{Error Distribution Smoothing: Advancing \\ Low-Dimensional Imbalanced Regression}



\icmlsetsymbol{equal}{*}

\begin{icmlauthorlist}
\icmlauthor{Donghe Chen}{buaa}
\icmlauthor{Jiaxuan Yue}{buaa}
\icmlauthor{Tengjie Zheng}{buaa}
\icmlauthor{Lanxuan Wang}{buaa}
\icmlauthor{Lin Cheng}{buaa}
\end{icmlauthorlist}

\icmlaffiliation{buaa}{School of Astronautics, Beihang University, 100191 Beijing, China}

\icmlcorrespondingauthor{Lin Cheng}{chenglin5580@buaa.edu.cn}

%

\icmlkeywords{Supervised Learning, Data Imbalance, Mechine Learning, Regression}

\vskip 0.3in
]



\printAffiliationsAndNotice{\icmlEqualContribution} 

\begin{abstract}
  In real-world regression tasks, datasets frequently exhibit imbalanced distributions, characterized by a scarcity of data in high-complexity regions and an abundance in low-complexity areas. This imbalance presents significant challenges for existing classification methods with clear class boundaries, while highlighting a scarcity of approaches specifically designed for imbalanced regression problems. To better address these issues, we introduce a novel concept of Imbalanced Regression, which takes into account both the complexity of the problem and the density of data points, extending beyond traditional definitions that focus only on data density. Furthermore, we propose Error Distribution Smoothing (EDS) as a solution to tackle imbalanced regression, effectively selecting a representative subset from the dataset to reduce redundancy while maintaining balance and representativeness. Through several experiments, EDS has shown its effectiveness, and the related code and dataset can be accessed at https://anonymous.4open.science/r/Error-Distribution-Smoothing-762F.
\end{abstract}

\section{Introduction}

In the realm of supervised learning, real-world datasets frequently exhibit skewed distributions, where certain regions of the outcome space contain significantly fewer samples \cite{zhang_deep_2023,ding_deep_2022}. This data imbalance can adversely affect the predictive performance of models, particularly for less represented outcomes, and has driven the development of various mitigation techniques \cite{song_learning_2022,chawla_smote_2011,he_adasyn_2008}. While much of the existing literature focuses on addressing imbalances in categorical outcomes \cite{buda_systematic_2018}, many practical applications involve continuous variables such as age or health metrics (e.g., heart rate, blood pressure). Discretizing continuous variables into categories neglects their inherent ordinal relationships and continuity, potentially leading to suboptimal model performance. Therefore, there is a need for specialized approaches that consider the unique characteristics of continuous targets to improve regression performance.

Imbalanced regression presents distinct challenges not fully addressed by methods designed for classification tasks and has received relatively limited attention in the literature \cite{yang_delving_2021}. Unlike classification, regression involves predicting continuous labels without clear class boundaries, complicating the direct application of traditional imbalance mitigation techniques such as resampling or reweighting. For example, in the prediction of house prices, each instance typically has a unique value, making it impractical to apply class-based methods directly. Existing solutions have adapted techniques like SMOTE for regression by generating synthetic samples through interpolation \cite{camacho_wsmoter_2024} or utilizing bagging-based ensemble methods \cite{branco_rebagg_2018}. However, these adaptations often fail to account for the varying complexity of the underlying functions and the differing sample requirements across different approaches.

We introduce a novel definition for imbalanced regression that incorporates both data density and regression complexity via the Complexity-to-Density Ratio (CDR). Unlike traditional definitions that focus solely on data density, it provides a more comprehensive framework for evaluating and addressing low-dimensional imbalanced regression issues. To effectively address the challenges of imbalanced regression, we propose Error Distribution Smoothing (EDS), which selectively reduces data pairs in overrepresented regions to eliminate redundancy and smooths the distribution of prediction errors across different regions. This approach ensures consistent model performance in underrepresented regions while maintaining accuracy in overrepresented regions. We also provide a set of diverse datasets for benchmarking to support the practical assessment of imbalanced regression solutions.

Our contributions are as follows:
\begin{itemize}
  \item A quantified definition of imbalanced regression is introduced, proposing the Complexity-to-Density Ratio (CDR) as a quantified metric for evaluating imbalanced regression.

  \item EDS finely quantifies and assesses redundancy, unlike other methods that reduce redundant data in a more coarse manner. By precisely evaluating and mitigating imbalance through smoothing prediction errors, it enhances prediction consistency and computational efficiency.
\end{itemize}

\section{Related Work}

In the field of imbalanced classification, solutions are generally categorized into data-centric and algorithmic approaches. Data-centric methods address class imbalance by altering dataset composition through techniques such as minority oversampling or majority undersampling \cite{chawla_smote_2011,he_adasyn_2008}, with SMOTE being a prominent technique for generating synthetic instances via linear interpolation \cite{chawla_smote_2011,zhang_mixup_2018,verma_manifold_2019}. Algorithmic methods enhance classifier performance by adjusting the learning process, including loss function weighting \cite{cao_learning_2019,cui_class-balanced_2019}, transfer learning \cite{singh_imbalanced_2021}, and meta-learning \cite{shu_meta-weight-net_2019}. Recent studies highlight that semi-supervised and self-supervised learning paradigms can significantly improve classification outcomes in imbalanced datasets \cite{yang_rethinking_2020}.

Imbalanced regression presents distinct challenges and has received relatively less attention in the literature compared to classification tasks. Adaptations from classification approaches have been applied, such as generating synthetic observations for sparsely populated target regions through interpolation \cite{camacho_wsmoter_2024} or by introducing Gaussian noise \cite{branco_smogn_2017}. Moreover, ensemble methods that integrate multiple preprocessing strategies have also been investigated \cite{branco_rebagg_2018}. Nonetheless, existing methods often fail to account for the varying complexity of regression across different regions, leading to differing data requirements that are not adequately addressed, an issue this paper resolves by proposing a novel approach to effectively capture and manage these regional complexities.

\section{Variable Definition}

To quantitatively evaluate imbalanced regression problems, we have developed two metrics alongside an optimization approach. The Complexity-to-Density Ratio (CDR) assesses whether regions within the data feature space contain a sufficient number of data pairs relative to their complexity. Meanwhile, the Global Imbalance Metric (GIM) provides an evaluation of the overall data distribution across all regions to identify any imbalances that could impact model performance. Additionally, we formulate an optimization problem aimed at selecting a representative subset from the dataset, ensuring the sample is both balanced and informative for training predictive models.

The mathematical notations for imbalanced regression are defined and the Complexity-to-Density Ratio (CDR) is introduced. Table \ref{tab:variables} outlines key symbols and their descriptions, with the notation for feature vectors, label vectors, datasets, and measures of region size and complexity being established.
\begin{table*}[htbp]
  \centering
  \vskip 0.15in 
  \caption{Mathematical Notations and Definitions for Imbalanced Regression}
  \label{tab:variables}
  \begin{small}
  \begin{tabular}{lll}
    \toprule
    \textbf{Symbol} & \textbf{Description} & \textbf{Mathematical Expression} \\
    \midrule
    $\mathcal{X}$ & n-dimension of the feature space & $\mathcal{X} \subseteq \mathcal{R}^n$\\
    $\mathcal{Y}$ & m-dimension of the label space & $\mathcal{Y}\subseteq \mathcal{R}^m$\\
    $\boldsymbol{x}_i$ & feature vector for the $i$-th data pair & $\boldsymbol{x}_i \in \mathcal{X}$ \\
    $\boldsymbol{y}_i$ & label vector for the $i$-th data pair & $\boldsymbol{y}_i \in \mathcal{Y}$ \\
    $\boldsymbol{y=f(x)}$ & Target function between feature and label &$f:\mathcal{X} \rightarrow \mathcal{Y}$\\
    $\boldsymbol{\hat{y}=\hat{f}(x)}$ & Linear interpolation models between feature and label &$\hat{f}:\mathcal{X} \rightarrow \mathcal{Y}$\\
    $\boldsymbol{e(x)}$ & Prediction error of the linear interpolation model$\hat{f}$ & $\boldsymbol{e(x) = \hat{f}(x) - f(x)}$\\
    $e_i$ & Prediction error for data pair $(\boldsymbol{x}_i, \boldsymbol{y}_i) \in \mathcal{D}$ & $e_i = \|\boldsymbol{\hat{f}(x) - f(x)}\|_2$ \\
    $\mathcal{D}$ & Entire Dataset with $N$ elements & $\mathcal{D} = \{(\boldsymbol{x}_i, \boldsymbol{y}_i)\}_{i=1}^{N}$ \\
    $\mathcal{D}_M$ & Minor Dataset (Randomly Sampled from $\mathcal{D}$) & $\mathcal{D}_M \subset \mathcal{D}$ \\
    $\mathcal{D}_R$ & Representative Dataset & $\mathcal{D}_R \subset \mathcal{D}$\\
    $\mathcal{D}_A$ & Auxiliary Dataset & $\mathcal{D}_A = \mathcal{D} \setminus \mathcal{D}_R$ \\
    $|\cdot|$ & Cardinality of a set, i.e., the number of elements in the set & \\
    $\Omega$ & A closed subset of the feature space & $\Omega \subset \mathcal{X}$\\
    $\lambda(\Omega)$ & Measure of the set $\Omega$ & $ \lambda(\Omega) = \int_{\boldsymbol{x} \in \Omega} d\boldsymbol{x}$\\
    $\mathcal{F}$ & The feature space is partitioned into a set of non-overlapping regions. & $\mathcal{F} = \{\Omega_i\}_{i=1}^{k},M(\Omega_i \cap \Omega_j) = 0$\\
    $\mathcal{H}(\boldsymbol{x})$ & Hessian tensor of the target function at $\boldsymbol{x}$ &  $\mathcal{H}(\boldsymbol{x})_{ijk} = \frac{\partial f\boldsymbol{(x)}_i}{\partial x_j \partial x_k}$ \\
    $\|\mathcal{H}(\boldsymbol{x})\|_F$ & Frobenius norm of the Hessian tensor at $\boldsymbol{x}$ & $\sqrt{\sum_{ijk} \mathcal{H}(\boldsymbol{x})_{ijk}^2}$ \\
    $\psi$ & A human-defined error threshold & $\psi > 0   $ \\
    \bottomrule
  \end{tabular}
\end{small}
\vskip -0.1in
\end{table*}

\begin{definition}[Complexity-to-Density Ratio (CDR)]

The Complexity-to-Density Ratio (CDR) for a region $\Omega$ is given by
\begin{equation}
  \rho(\Omega,\mathcal{D}) = \frac{g_c(\Omega) }{|\Omega \cap \mathcal{D}| / g_s(\Omega)} =  \frac{g_c(\Omega) \cdot g_s(\Omega)}{|\Omega \cap \mathcal{D}|}
\end{equation}
where $ |\Omega \cap \mathcal{D}| $ denotes the number of data pairs in $\mathcal{D}$ that lie within $\Omega$. The functions $ g_c(\cdot) $ and $ g_s(\cdot) $ can be defined as follows:

\begin{itemize}
  \item $g_c(\Omega)$ measures complexity via the maximum Frobenius norm of the Hessian tensor across $\Omega$:
  \begin{equation}
    g_c(\Omega) = \max_{\boldsymbol{x} \in \Omega} \|\boldsymbol{H}(\boldsymbol{x})\|_F
  \end{equation}
  \item $g_s(\Omega)$ quantifies the region's size through the maximum squared Euclidean distance between any two points in $\Omega$:
  \begin{equation}
    g_s(\Omega) = \max_{\boldsymbol{x}_1, \boldsymbol{x}_2 \in \Omega} \|\boldsymbol{x}_1 - \boldsymbol{x}_2\|_2^2
  \end{equation}
\end{itemize}

Note that $ g_c(\cdot) $ and $ g_s(\cdot) $ are tailored for our specific context and can be adjusted based on the dataset and task. Their definitions should align with the problem characteristics and desired CDR properties.
\end{definition}

To quantify the distribution of complexity-to-density ratios across all regions $\mathcal{F} = \{\Omega_i\}_{i=1}^{k}$, we define the \textbf{Log-CDR distribution} ($N(\mu,\sigma^2)$). It models the log-transformed CDR as a normal distribution, summarizing dataset imbalance.

\begin{definition}[Log-CDR distribution]
The Log-CDR distribution $I(\mathcal{F},\mathcal{D})=N(\mu,\sigma^2)$ is constructed by estimating the parameters $\hat{\mu}$and $\hat{\sigma}^2$ of the underlying normal distribution from the transformed values $\{\ln(\rho(\Omega_j, \mathcal{D}))\}_{j=1}^{k}$ using Maximum Likelihood Estimation (MLE):
\begin{equation}
  \begin{aligned}
    \hat{\mu} &= \frac{1}{k}\sum_{j=1}^{k} \ln(\rho(\Omega_j, \mathcal{D})) \\
    \hat{\sigma}^2 &= \frac{1}{k-1}\sum_{j=1}^{k} (\ln(\rho(\Omega_j, \mathcal{D})) - \hat{\mu})^2
    \end{aligned}
\end{equation}
\end{definition}
Regions are classified into High-CDR, Medium-CDR, and Low-CDR categories based on their complexity-to-density ratio ($\rho(\Omega_j, \mathcal{D})$) relative to the estimated mean ($\hat{\mu}$) and standard deviation ($\hat{\sigma}$), with $z>0$ indicating the number of standard deviations from the mean. The classification is determined by the following inequalities:
\begin{equation}
  \Omega = 
  \begin{cases}
    \text{High-CDR}, & \text{if } \ln(\rho) > \hat{\mu} + z\hat{\sigma} \\
    \text{Medium-CDR}, & \text{if } \hat{\mu} - z\hat{\sigma} \leq \ln(\rho) \leq \hat{\mu} + z\hat{\sigma} \\
    \text{Low-CDR}, & \text{if } \ln(\rho) < \hat{\mu} - z\hat{\sigma}
  \end{cases}
\end{equation}

\section{Problem Formulation and Reconstrunction}
The optimization problem is formulated to select the smallest possible representative subset $\mathcal{D}_R$ from the entire dataset $\mathcal{D}$, ensuring that strict complexity-to-sample density ratio (CDR) constraints are adhered to while minimizing the number of samples. This approach aims to mitigate dataset imbalance and manage risk at a specified confidence level. The optimization problem is defined as follows:
\begin{equation}
  \begin{aligned}
  & \underset{\mathcal{D}_R \subseteq \mathcal{D}}{\text{minimize}}
  & & |\mathcal{D}_R| \\
  & \text{subject to}
  & & \hat{\mu}|_{I(\mathcal{F}, \mathcal{D}_R)} + z\hat{\sigma}|_{I(\mathcal{F}, \mathcal{D}_R)} \leq \psi\\
  \end{aligned}
\end{equation}
where $\hat{\mu}$ and $\hat{\sigma}$ denote the mean and standard deviation of the log-transformed CDR in $\mathcal{D}_R$. The constraint keeps the log-transformed CDR below a threshold $\psi$ at a given confidence level, with $z$ indicating standard deviations from the mean. Minimizing $|\mathcal{D}_R|$ while preserving representativeness enhances efficiency and reliability in data processing, particularly in resource-limited settings or for rapid insights.

In regression analysis, the lack of global information about the true function $f(x)$ renders complexity assessment methods based on derivatives or higher-order derivatives infeasible. To address this challenge, we propose a method that combines Delaunay Triangulation with linear interpolation models to approximate the CDR in a complexly partitioned feature space.To facilitate the optimization process, we introduce three critical assumption:

\begin{assumption}[Discretization of Evaluation Metrics]
  \label{assump:discretization}
  Let the domain of evaluation metrics be potentially continuous and infinite, posing computational and mathematical challenges. We assume that the evaluation process can be constrained to a finite set of discrete samples. Given an original dataset $ D = \{(\boldsymbol{x}_i, y_i)\}_{i=1}^{N} $ where $\boldsymbol{x}_i \in \mathbb{R}^n$ is a feature vector and $y_i \in \mathbb{R}^m$ is a label for each data pair, discretization involves selecting a subset $ D_R \subseteq D $ such that $ |D_R| \leq |D| $.
\end{assumption}

\begin{assumption}[Finite Partitioning of Feature Space]
  \label{assump:partitioning}
  We assume it is feasible to extract a finite subset of samples from the feature space $X$ and apply Delaunay triangulation to divide $X$ into a set of non-overlapping simplices $\mathcal{F} = \{\Omega_j\}_{j=1}^{k}$. Each simplex $\Omega_j$ represents a distinct local region within $X$, ensuring no overlap between any two simplices, i.e., $M(\Omega_i \cap \Omega_j) = 0$ for all $i \neq j$.
\end{assumption}

\begin{assumption}[Simplified Constraint Formulation]
  \label{assump:constraint}
  To mitigate computational intensity and mathematical complexity associated with estimating complex distributions or computing high-order derivatives, we simplify constraint formulation. Specifically, we require that the maximum Euclidean norm of the error vectors $ \|\boldsymbol{e}_i\|_2 $ satisfies $ \|\boldsymbol{e}_i\|_2 < \psi $ for predefined threshold $ \psi > 0 $. 
\end{assumption}

The proposed methodological choices simplify complex problems by discretizing evaluation metrics and partitioning the feature space, thereby reducing computational complexity and enabling accurate modeling with local linear interpolation. Assumptions \ref{assump:discretization} and \ref{assump:partitioning} are intuitively reasonable, while Assumption \ref{assump:constraint} requires validation through analyzing the correlation between error and CDR. 

Our analysis reveals that significant errors predominantly occur in regions with maximal CDR, enabling effective management of the upper error bound within any region $\Omega$, as detailed in Appendix.~\ref{appendix:LIMs}. The upper bound on the maximum error within any region $\Omega$ can be expressed as follows, with a detailed derivation provided in Appendix.~\ref{appendix:LIMs}.

\begin{equation}
  \label{eq:error_bound}
  \begin{aligned}
    \ub_{\boldsymbol{x} \in \Omega}\|\boldsymbol{e(x)}\|_2
    &= \frac{1}{2} \max_{\boldsymbol{x}_1, \boldsymbol{x}_2 \in \Omega} \|\boldsymbol{x}_1 - \boldsymbol{x}_2\|_2^2  \max_{\boldsymbol{x} \in \Omega}\|\boldsymbol{H(x)}\|_F  \\
    &= \frac{1}{2} g_s(\Omega)  g_c(\Omega) \\
    &= \frac{g_c(\Omega) g_s(\Omega)}{|\Omega \cap \mathcal{D}|}  \frac{|\Omega \cap \mathcal{D}|}{2} 
  \end{aligned}
\end{equation}
where $\ub_{\boldsymbol{x} \in \Omega}\|\boldsymbol{e(x)}\|_2$ denotes the upper bound on the maximum error within region $\Omega$. This expression shows that the upper bound is proportional to both the size of the region $g_s(\Omega)$ and its complexity $g_c(\Omega)$, normalized by the number of data points within the region.

When using Delaunay triangulation, the number of vertices in an $n$-dimensional simplex remains constant ($|\Omega \cap \mathcal{D}| = n + 1$). Consequently, this simplifies the computation of the upper bound. The proportionality of the maximum error within any region $\Omega$ to the complexity-to-density ratio (CDR) is given by:

\begin{equation}
  \ub_{\boldsymbol{x} \in \Omega}\|\boldsymbol{e}(\boldsymbol{x})\|_2 \propto \rho(\Omega, \mathcal{D})
\end{equation}

where $\rho(\Omega, \mathcal{D})$, the CDR within a region, combines the effects of $g_c(\Omega)$ (maximum complexity) and $g_s(\Omega)$ (region size). This relationship highlights how the maximum error scales with the characteristics of region $\Omega$. Specifically, regions with higher complexity or larger size will exhibit greater potential for error, while regions with more data points tend to have lower errors due to better representation.

The upper bound of error within any region $\Omega \in \mathcal{F}$ is related to the statistical properties of the data. Specifically, at a 98.85\% confidence level, the threshold $\hat{\mu} + 2\hat{\sigma}$ captures nearly all significant error values. This implies that the maximum error $\max_{\boldsymbol{x} \in \Omega} \|\boldsymbol{e}\|_2$ can be considered equivalent to this threshold, with an extremely low probability (less than 1.15\%) of observing errors exceeding this limit. Therefore, the relationship between the upper bound of error and the statistical properties is expressed as:

\begin{equation}
  \max_{\Omega \in \mathcal{F}} \ub_{\boldsymbol{x} \in \Omega} \|\boldsymbol{e}(\boldsymbol{x})\|_2 \propto \hat{\mu}|_{I(\mathcal{D}_R)} + 2\hat{\sigma}|_{I(\mathcal{D}_R)}
\end{equation}

The following assumptions underpin our approach: errors $\|\boldsymbol{e}\|_2$ within each region $\Omega$ are assumed to approximately follow a log normal distribution; the variance of errors is considered relatively consistent across different regions (homogeneity of variance); and a sufficiently large sample size ensures reliable estimation of $\hat{\mu}$ and $\hat{\sigma}$.The 98.85\% confidence level provides a conservative estimate capturing almost all significant errors, accounting for potential non-normality in the data. The threshold $\hat{\mu} + 2\hat{\sigma}$ serves as a robust upper bound, indicating that most observed errors will not exceed this value. This approximation is highly reliable, with only a minimal probability of encountering larger errors.

\section{Representative Dataset Construction}

To construct the representative dataset $\mathcal{D}_R$, we employ Delaunay triangulation with Linear Interpolation Models (LIMs)\cite{chang_polynomial_2018,chang_leveraging_2025,berrut_barycentric_2004}. This method tessellates the feature space into non-overlapping simplices, ensuring that no data pairs lie within a simplex's circumcircle\cite{boissonnat_incremental_2009}. This approach accurately captures local structures  while preventing overfitting. By constructing simplices using a representative dataset, the method preserves computational efficiency even for large datasets.

\begin{figure}[htbp]
  \centering
  \includegraphics[width=0.35\textwidth]{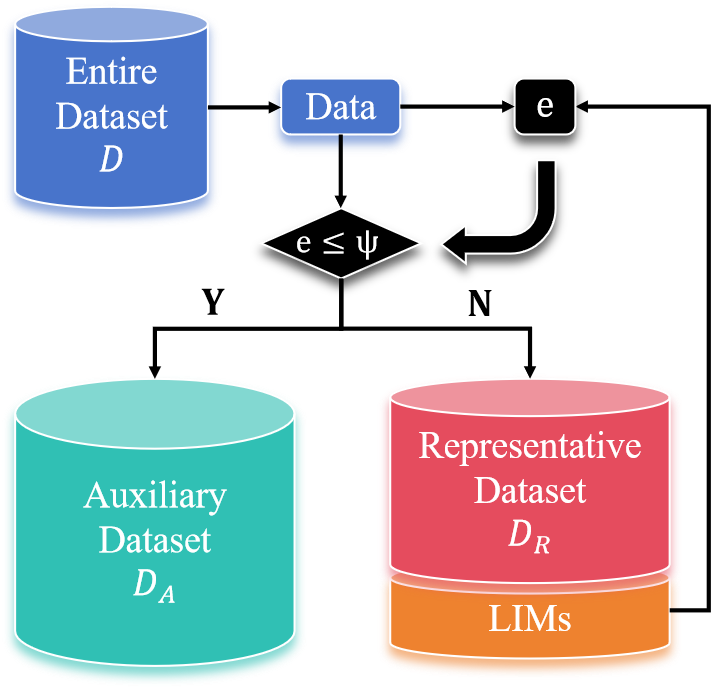}
  \caption{Flowchart illustrating the procedure for generating the Representative Dataset by allocating data to either the Representative Dataset or the Auxiliary Dataset based on prediction error ($e$) relative to a defined threshold ($\psi$).}
  \vspace{0.1in}
  \label{fig:data_allocation_flowchart}
  \vskip -0.1in
\end{figure}

Figure \ref{fig:data_allocation_flowchart} provides a visual guide to the process of constructing the representative dataset $\mathcal{D}_R$ using EDS and generating LIMs.Data allocation between $\mathcal{D}_R$ and  $\mathcal{D}_A$ is based on prediction error relative to a threshold $\psi$. Data pairs with errors less than $\psi$ are assigned to $\mathcal{D}_A$, while those with higher errors are included in $\mathcal{D}_R$. Linear interpolation models constructed using $\mathcal{D}_R$ then compute prediction errors and support further analysis.

\begin{algorithm}[htbp]
  \caption{Constructing $\mathcal{D}_R$ and $\mathcal{D}_A$}
  \label{alg:dataProcessing}
  \renewcommand{\algorithmicrequire}{\textbf{Input:}}
  \renewcommand{\algorithmicensure}{\textbf{Output:}}
  \begin{algorithmic}[1]
    \REQUIRE 
        $\mathcal{D} = \{ (\boldsymbol{x}_i, \boldsymbol{y}_i) \}_{i=1}^{N}$: Entire Dataset with feature vectors $\boldsymbol{x}_i$ and label vectors $\boldsymbol{y}_i$ \\
        $B$: Batch size  \\
        $\psi$: Error threshold 

    \ENSURE Representative Dataset $\mathcal{D}_R$ and Auxiliary Dataset $\mathcal{D}_A$

    \STATE Initialize: Randomly select $n+1$ instances from $\mathcal{D}$ to form $\mathcal{D}_R$; set $\mathcal{D}_A \leftarrow \emptyset$
    \STATE Perform initial Delaunay triangulation on $\mathcal{D}_R$ to obtain $\mathcal{T}$
    
    \FOR{each batch of data pairs $\mathcal{B} \subseteq \mathcal{D}$ of size $B$}
      \FOR{each data pair $(\boldsymbol{x}_i, \boldsymbol{y}_i) \in \mathcal{B}$}
        \STATE Determine which simplex $\mathcal{S}_j \in \mathcal{T}$ contains $\boldsymbol{x}_i$
        \IF{$\boldsymbol{x}_i$ is not in any $\mathcal{S}_j$}
          \STATE Add $\boldsymbol{x}_i$ to $\mathcal{D}_R$, update $\mathcal{T}$ by inserting $\boldsymbol{x}_i$ into the triangulation
        \ELSE
          \STATE Predict labels using the model, calculate the prediction error
          \IF{prediction error $> \psi$}
            \STATE Add $\boldsymbol{x}_i$ to $\mathcal{D}_A$
          \ENDIF
        \ENDIF
      \ENDFOR
    \ENDFOR
    
    \STATE \textbf{return}  $\mathcal{D}_R$ and $\mathcal{D}_A$
  \end{algorithmic}
\end{algorithm}

The feature space is partitioned into simplices, with Linear Interpolation Models (LIMs) built from samples at the vertices. Error Distribution Smoothing (EDS) adjusts sample density based on the local Complexity-to-Density Ratio (CDR) to ensure balanced representation. This method retains critical data points in high-CDR regions and avoids oversampling in low-CDR areas. Algorithm~\ref{alg:dataProcessing} constructs the representative dataset $\mathcal{D}_R$ and auxiliary dataset $\mathcal{D}_A$, ensuring that only essential points are kept in $\mathcal{D}_R$. The output is the updated sets $\mathcal{D}_R$ and $\mathcal{D}_A$.

\section{Experiment}

This section evaluates Error Distribution Smoothing (EDS), which improves model accuracy and generalization. EDS enhances prediction accuracy in systems like the Lorenz system, reduces dimensionality in high-dimensional data, and improves real-world applications such as quadcopter dynamics by lowering prediction errors and training time.

\subsection{Motivation Example}

To illustrate Error Distribution Smoothing (EDS), we use the function $ f(x_1, x_2) = (0.33 + x_1^2 + x_2^2)^{-1} $, where $ x_1 $ and $ x_2 $ are uniformly sampled from $[-3, 3]$. This function shows rapid changes near the $(0,0)$ due to the quadratic terms in the denominator, with variations decreasing as points move away from $(0,0)$. Accurately capturing this behavior, particularly the high complexity near the origin, underscores the need for an adaptive sampling strategy.

\begin{figure}[htbp]
  \centering
  \includegraphics[width=0.48\textwidth]{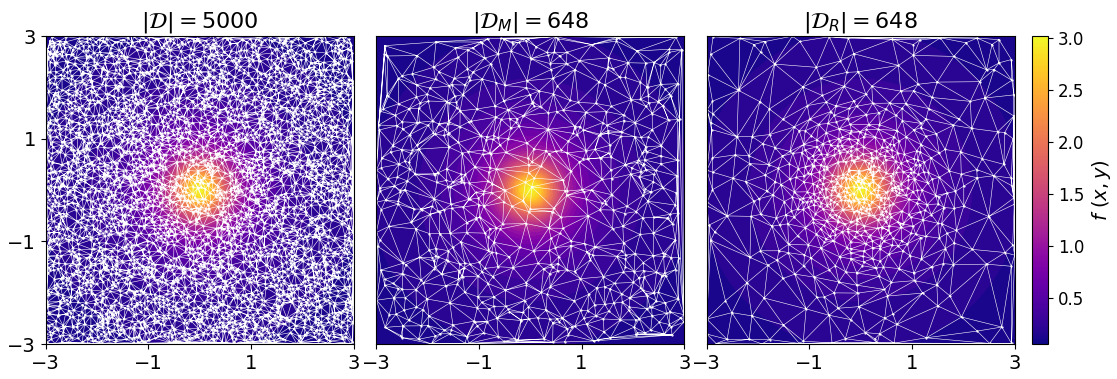}
  \caption{The Delaunay triangulation for datasets $\mathcal{D}$, $\mathcal{D}_M$, and $\mathcal{D}_R$ reveals uneven partitioning characteristics, with dense triangulations near the origin highlighting areas of significant function variation, particularly pronounced in $\mathcal{D}_R$.}
  \label{fig:simplices}
  \vskip -0.1in
\end{figure}

As shown in Figure \ref{fig:simplices}, the Delaunay triangulations reveal the non-uniformity of the original data space partitions. By focusing on areas with higher function variability, especially around the origin, this figure underscores the importance of adaptive sampling strategies that can lead to more effective learning processes.

\begin{figure}[htbp]
  \centering
  \includegraphics[width=0.48\textwidth]{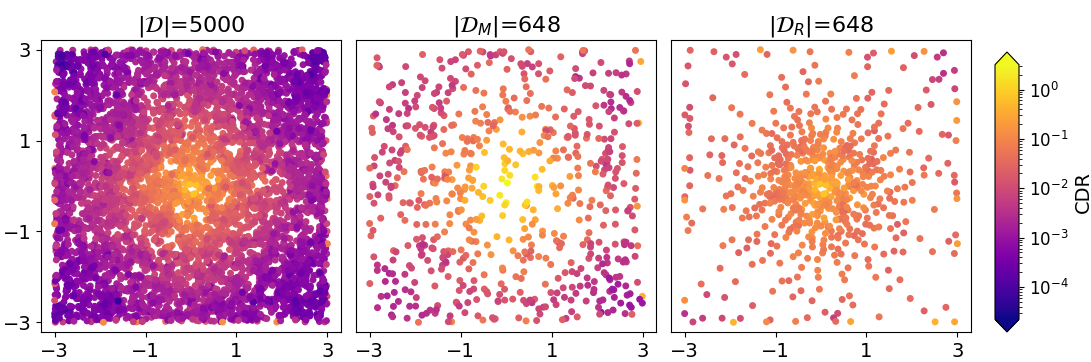}
  \caption{A comparison of conditional density ratios (CDRs) for datasets $\mathcal{D}$, $\mathcal{D}_M$, and $\mathcal{D}_R$ shows that $\mathcal{D}_R$ exhibits a more uniform CDR distribution, which is indicative of effective error distribution smoothing through adjusted sample density corresponding to regional complexity.}
  \label{fig:scatter_CDR}
  \vskip -0.1in
\end{figure}

Figure \ref{fig:scatter_CDR} offers insights into the comparative consistency of conditional density ratios (CDRs) across different datasets. It is clear that $\mathcal{D}_R$ benefits from an optimized sample allocation that corresponds to the underlying complexity of the data region, leading to a smoother error distribution.

\begin{figure}[htbp]
  \centering
  \includegraphics[width=0.48\textwidth]{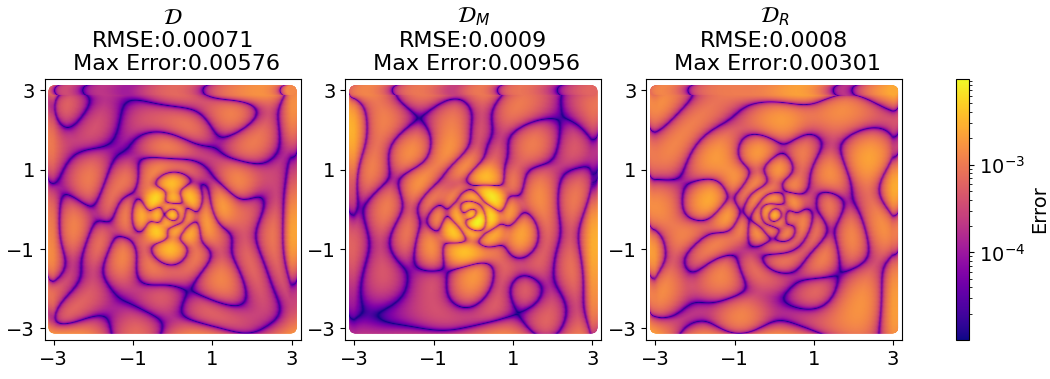}
  \caption{Performance evaluation of a MLP trained on datasets $\mathcal{D}$, $\mathcal{D}_M$, and $\mathcal{D}_R$ indicates that the MLP trained on $\mathcal{D}_R$ achieves a more uniform error distribution, underscoring the advantages of EDS in improving model generalization and robustness.}
  \label{fig:valid_MLP}
  \vskip -0.1in
\end{figure}
In Figure \ref{fig:valid_MLP}, we observe the performance improvements of a multi-layer perceptron (MLP) when it is trained on the $\mathcal{D}_R$ dataset. The increased accuracy and reduced variance in prediction errors emphasize the role of EDS in achieving better generalization and robustness of regression models.

Collectively, Figures \ref{fig:simplices} to \ref{fig:valid_MLP} demonstrate that EDS effectively adjusts sample density according to regional complexity, enhancing CDR uniformity and model performance. These results highlight EDS's potential for improving the generalization and robustness of regression models in imbalanced datasets.

\subsection{Evaluating EDS in Dynamics System Identification}

To evaluate Error Distribution Smoothing (EDS) in the context of dynamic system identification, we applied the Sparse Identification of Nonlinear Dynamics (SINDY) algorithm \cite{brunton_discovering_2016} to the well-known Lorenz system. The Lorenz system, originally developed as a simplified mathematical model for atmospheric convection, is described by a set of three nonlinear differential equations that exhibit chaotic behavior under certain parameter settings. It is defined by Equation~\ref{eq:lorenz} with parameters $\sigma = 10$, $\rho = 28$, and $\beta = \frac{8}{3}$:

\begin{equation}
  \label{eq:lorenz}
  \begin{cases}
    \dot{x} = \sigma(y - x) \\
    \dot{y} = x(\rho - z) - y \\
    \dot{z} = xy - \beta z
  \end{cases}
\end{equation}
where the features consist of measurements of the state variables $x$, $y$, and $z$. The labels are the rates of change of these state variables, $\dot{x}$, $\dot{y}$, and $\dot{z}$. By applying SINDY, which identifies a sparse set of terms that best describe the dynamics from data, we aim to assess improvements in data representation and model performance through optimized sample distribution.

\begin{figure}[htbp]
  \centering
  \includegraphics[width=0.48\textwidth]{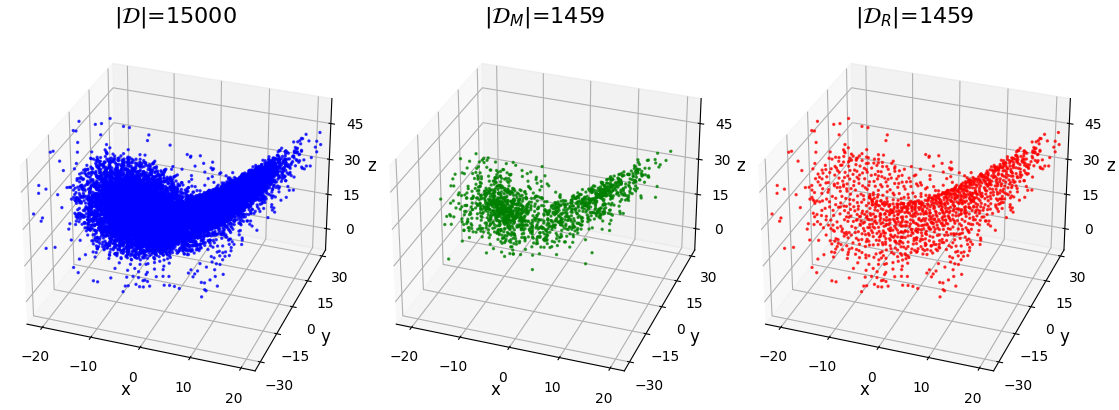}
  \vspace{0.1in}
  \caption{Comparison of Data Coverage: $\mathcal{D}$, $\mathcal{D}_M$, and $\mathcal{D}_R$. Despite fewer data pairs, $\mathcal{D}_R$ exhibits broad coverage, highlighting EDS's efficiency.}
  \label{fig:scatter_lorenz}
  \vspace{-0.1in}
\end{figure}

\begin{table}[htbp]
  \centering
  \vspace{0.1in}
  \caption{Comparison of SINDY performance}
  \label{tab:metrics}
  \renewcommand{\arraystretch}{2.0} 
  \setlength{\tabcolsep}{6pt} 
  \footnotesize 
  \begin{tabular}{lccc}
    \toprule
    \textbf{Item} & \textbf{$\mathcal{D}$} & \textbf{$\mathcal{D}_M$} & \textbf{$\mathcal{D}_R$} \\
    \midrule
    \textbf{$\Delta \dot{x}/10^{-2}$} & 
    $3x - 2y$ & 
    \makecell{$5x - 3y +$\\$0.1xz$} & 
    $x - 1.5y$ \\
    
    \textbf{$\Delta \dot{y}/10^{-2}$} & 
    \makecell{$-3x + 2y +$\\$0.1xz$} & 
    \makecell{$-5x + 3y +$\\$0.1xz$} & 
    $-x + 0.5y$ \\
    
    \textbf{$\Delta \dot{z}/10^{-2}$} & 
    $-0.2z$ & 
    $-0.2z$ & 
    $-0.1z$ \\
    
    RMSE & 0.0296 & 0.0485 & 0.0117 \\
    
    Max error & 0.715 & 1.161 & 0.189 \\
    
    Train time & 9.412 s & 0.058 s & 0.017 s \\
    \bottomrule
  \end{tabular}
  \vskip -0.1in
\end{table}

\begin{figure}[htbp]
  \centering
  \includegraphics[width=0.45\textwidth]{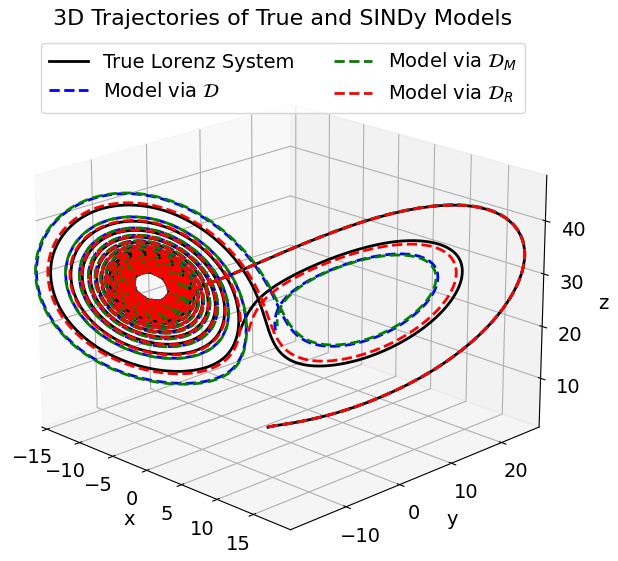}
  \vspace{0.1in}
  \caption{SINDY Forward Prediction Performance: Model via $\mathcal{D}_R$ (red) closely aligns with the true Lorenz trajectory (black), demonstrating superior precision.}
  \label{fig:sindy_predict}
  \vspace{-0.1in}
\end{figure}

Figure \ref{fig:scatter_lorenz} shows that $\mathcal{D}_R$, despite fewer samples, achieves broad coverage, improving the uniformity of the Complexity-to-Density Ratio (CDR). This leads to superior precision, as seen in Figure \ref{fig:sindy_predict}, where SINDY predictions using $\mathcal{D}_R$ closely match the true Lorenz trajectory. Table \ref{tab:metrics} further confirms that $\mathcal{D}_R$ results in lower RMSE, lower maximum error, and shorter training time. These findings demonstrate EDS's significant enhancement of data representation and model performance for system identification with SINDY on the Lorenz system.

\subsection{Evaluating EDS in High-Dimensional Regression}
To evaluate EDS in high-dimensional settings, we used a synthetic dataset of white rectangles on a black background, with each rectangle labeled by its polar moment of inertia for regression. To address the high-dimensional challenge, we extracted the positions of the rectangle edges to reduce dimensionality while preserving critical information, and subsequently standardized the features and labels to meet the model's requirements.

\begin{figure}[htbp]
  \centering
  \includegraphics[width=0.48\textwidth]{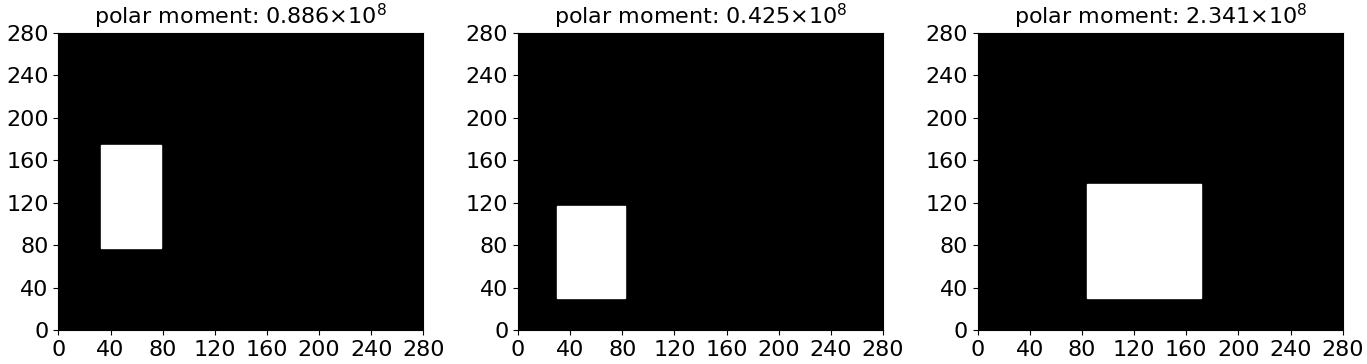}
  \caption{Multiple instances of a white rectangle on a black background, are labeled with the polar moment of inertia of the rectangle.}
  \label{fig:instances_rectangle}
  \vskip -0.1in
\end{figure}

Figure~\ref{fig:instances_rectangle} illustrates multiple instances of white rectangles on a black background, each labeled with its polar moment of inertia. This setup enables us to explore how well EDS can handle high-dimensional data when reduced to a meaningful lower-dimensional representation for regression. By focusing on these low-dimensional features, we ensure that critical information is retained, allowing EDS to function effectively.

\begin{figure}[htbp]
  \centering
  \includegraphics[width=0.48\textwidth]{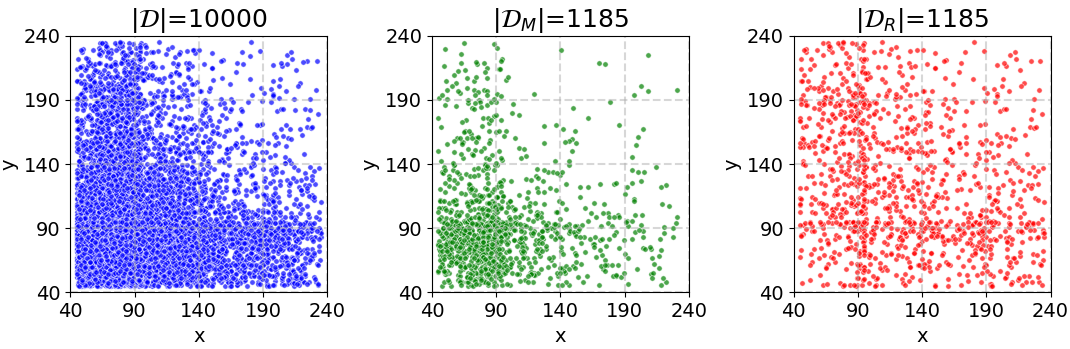}
  \caption{Distribution of Rectangle Centers in $\mathcal{D}$, $\mathcal{D}_M$, and $\mathcal{D}_R$. The spatial distribution is relatively uniform across datasets, with the highest data density observed in the lower left corner of the original image. Notably, the representative dataset $\mathcal{D}_R$ exhibits a more even spread, with reduced data density in its lower left corner compared to $\mathcal{D}$.}
  \label{fig:scatter_rectangle}
  \vskip -0.1in
\end{figure}

\begin{figure}[htbp]
  \centering
  \includegraphics[width=0.48\textwidth]{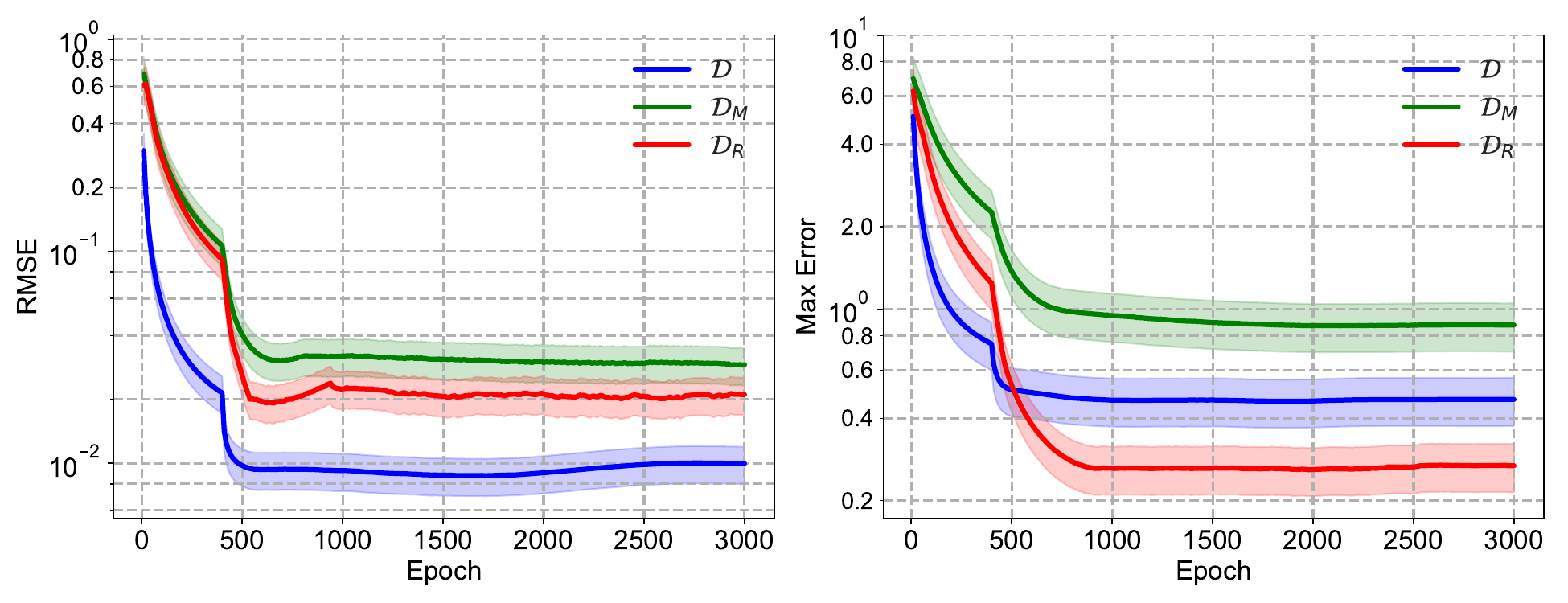}
  \caption{Comparison of MLP training on dimensionally reduced versions of $\mathcal{D}$, $\mathcal{D}_M$, and $\mathcal{D}_R$, evaluated on the Test Dataset.The MLP trained on $\mathcal{D}_R$, although having a higher RMSE, shows a significantly smaller Maximum Error, indicating enhanced robustness in worst-case scenarios.}
  \label{fig:MLP_rectangle}
  \vskip -0.1in
\end{figure}

Figures \ref{fig:scatter_rectangle} and \ref{fig:MLP_rectangle} illustrate how EDS optimizes sample distribution and boosts model performance. Figure \ref{fig:scatter_rectangle} shows that EDS balances the spatial distribution of rectangle centers across datasets $\mathcal{D}$, $\mathcal{D}_M$, and $\mathcal{D}_R$, improving data representation in lower dimensions. Figure \ref{fig:MLP_rectangle} demonstrates that MLP models trained on $\mathcal{D}_R$ achieve superior robustness and performance, particularly in challenging scenarios. By combining EDS with low-dimensional feature extraction, we effectively mitigate  dataset imbalance, enhancing both model performance and efficiency.

\subsection{Evaluating EDS in Real-World Problems}
\begin{figure}[htbp]
  \centering
  \begin{subfigure}[b]{0.20\textwidth}
      \includegraphics[width=\textwidth]{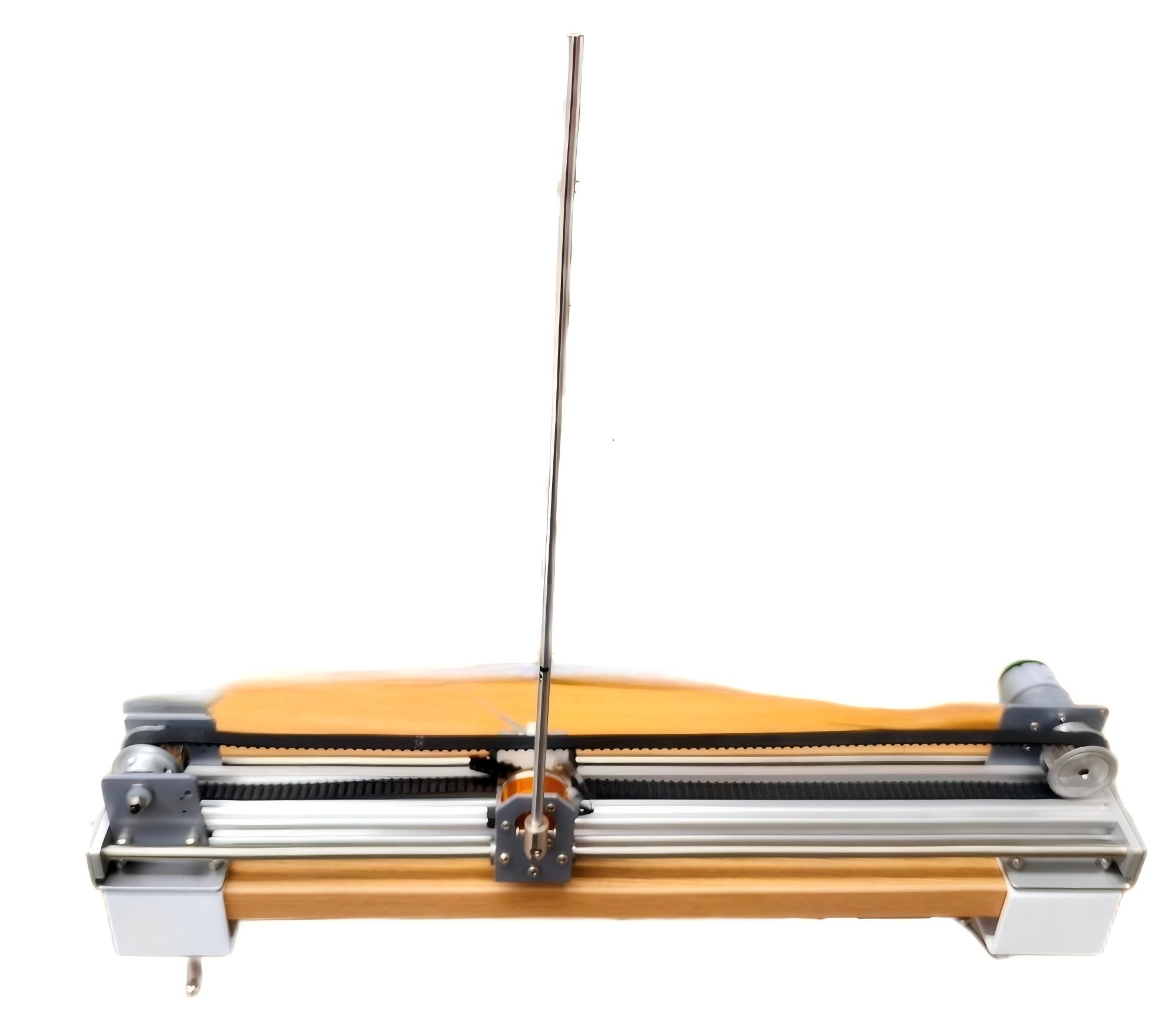}
      \caption{Cartpole}
      \label{fig:cartpole}
  \end{subfigure}
  \hfill
  \begin{subfigure}[b]{0.20\textwidth}
      \includegraphics[width=\textwidth]{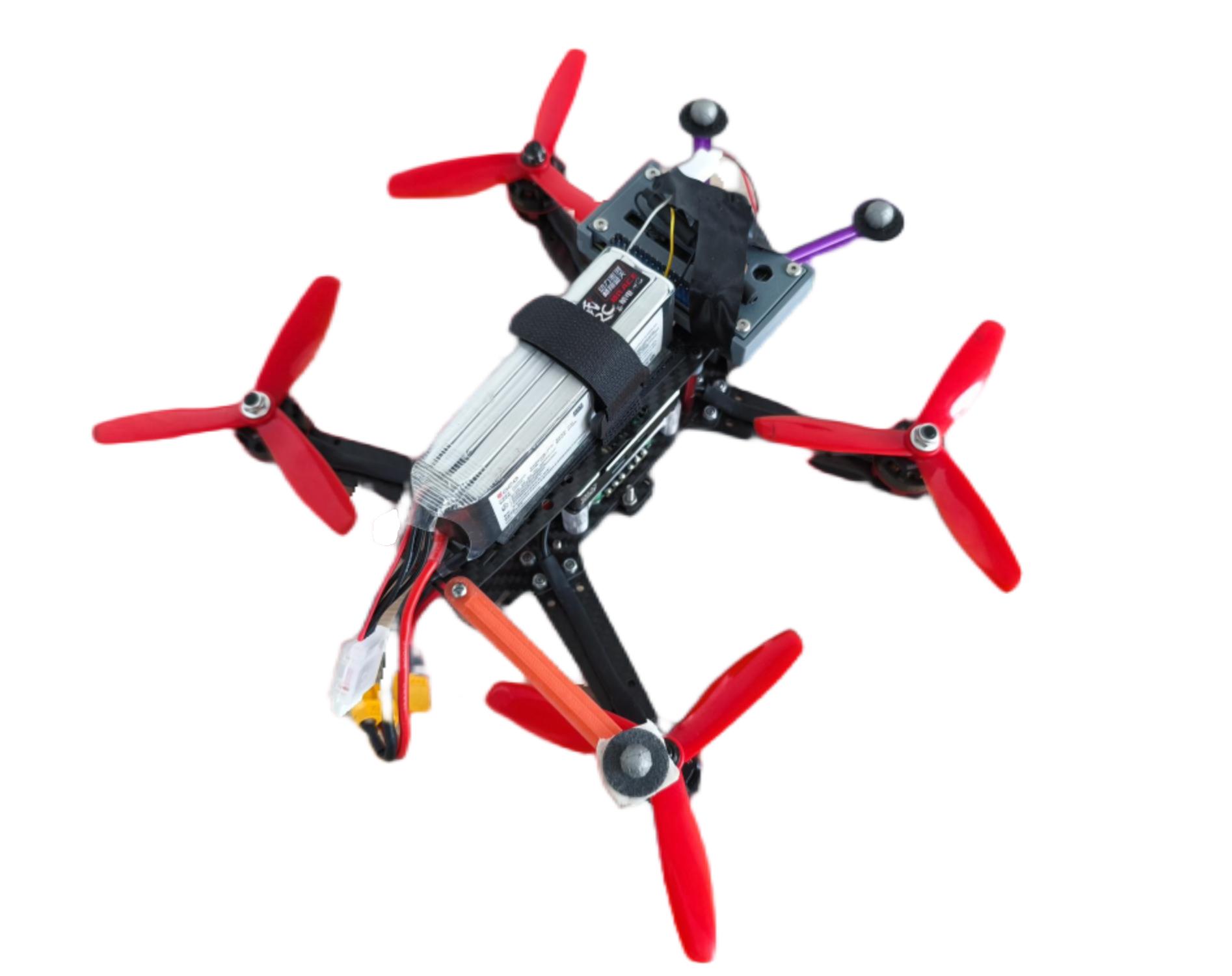}
      \caption{Quadcopter}
      \label{fig:quadcopter}
  \end{subfigure}
  \caption{Physical setups for dynamic systems used to evaluate control algorithms.}
  \label{fig:real_system_pic}
\end{figure}

To evaluate Error Distribution Smoothing (EDS) in handling real-world problems with strong noise and significant imbalance, we conducted experiments using two dynamic systems: the \textbf{Cartpole} and the \textbf{Quadcopter}. Data for both systems was collected under challenging conditions using PID controllers. For the Cartpole, features included the pole's angle, pole's angular velocity and motor's current, with labels being the cart's and pole's accelerations. For the Quadcopter, features included height, velocity and throttle, with labels being the acceleration. Figures~\ref{fig:cartpole} and~\ref{fig:quadcopter} depict the experimental setups.

\begin{figure}[htbp]
  \centering
  \begin{subfigure}[b]{0.48\textwidth}
      \includegraphics[width=\textwidth]{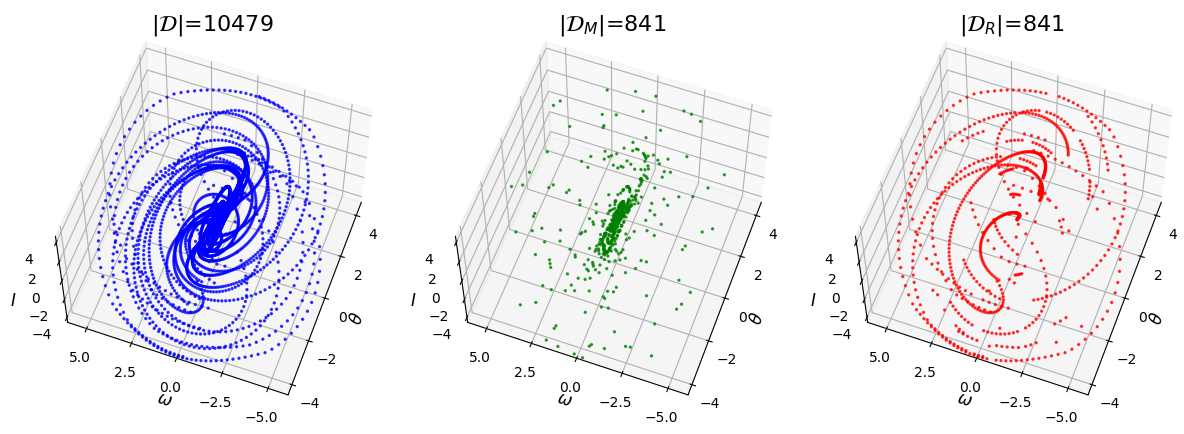}
      \caption{Cartpole}
      \label{fig:cartpole_dist}
  \end{subfigure}
  \hfill
  \begin{subfigure}[b]{0.48\textwidth}
      \includegraphics[width=\textwidth]{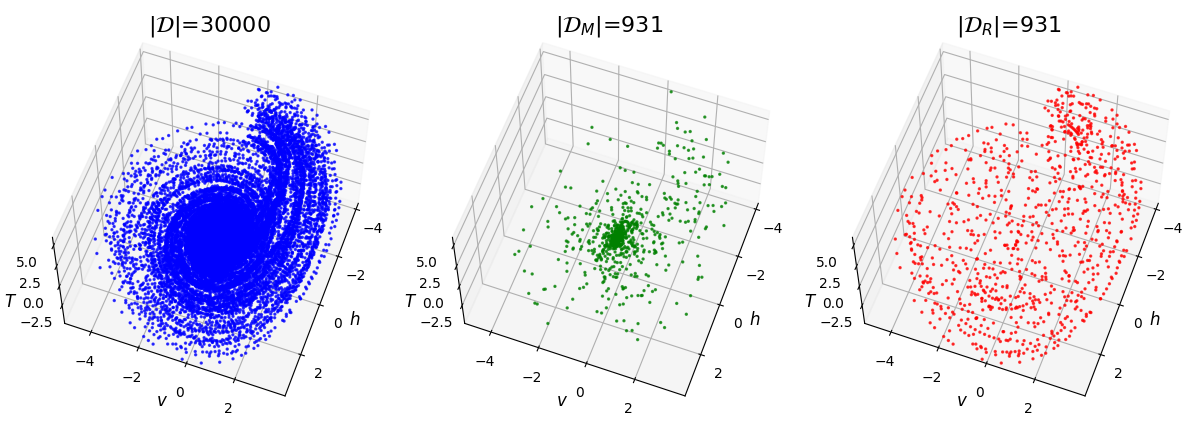}
      \caption{Quadcopter}
      \label{fig:quadcopter_dist}
  \end{subfigure}
  \caption{Comparison of Data Coverage Across Different Environments (Cartpole and Quadcopter): $\mathcal{D}$, $\mathcal{D}_M$, and $\mathcal{D}_R$.}
  \label{fig:real_system_dist}
  \vskip -0.1in
\end{figure}

Figures~\ref{fig:cartpole_dist} and~\ref{fig:quadcopter_dist} present a comparison of data coverage across different environments (Cartpole and Quadcopter). Despite having fewer data pairs, the $\mathcal{D}_R$ dataset demonstrates broad coverage, underscoring the efficacy of EDS in achieving comprehensive feature space exploration with reduced dataset sizes. This suggests that EDS can efficiently capture the essential characteristics of the state space even when the amount of available data is limited.

\begin{figure}[htbp]
  \centering
  \begin{subfigure}[b]{0.48\textwidth}
      \includegraphics[width=\textwidth]{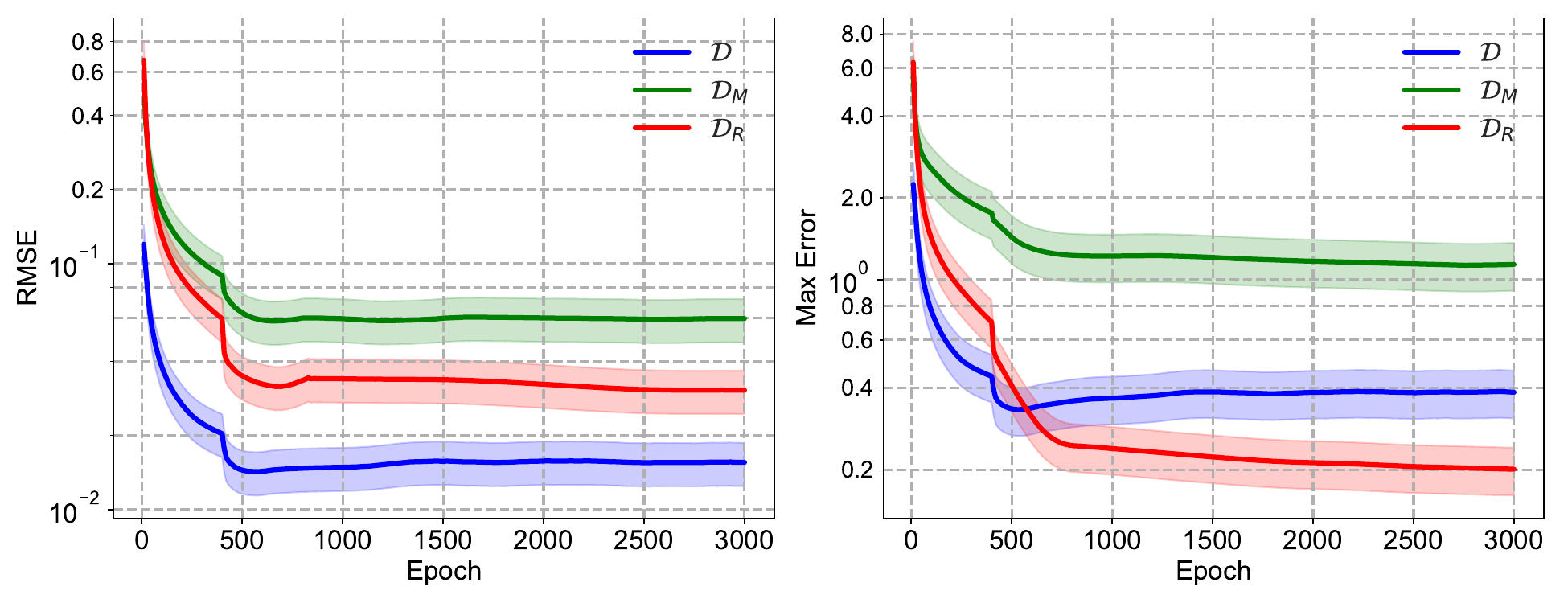}
      \caption{Cartpole}
      \label{fig:cartpole_error}
  \end{subfigure}
  \hfill
  \begin{subfigure}[b]{0.48\textwidth}
      \includegraphics[width=\textwidth]{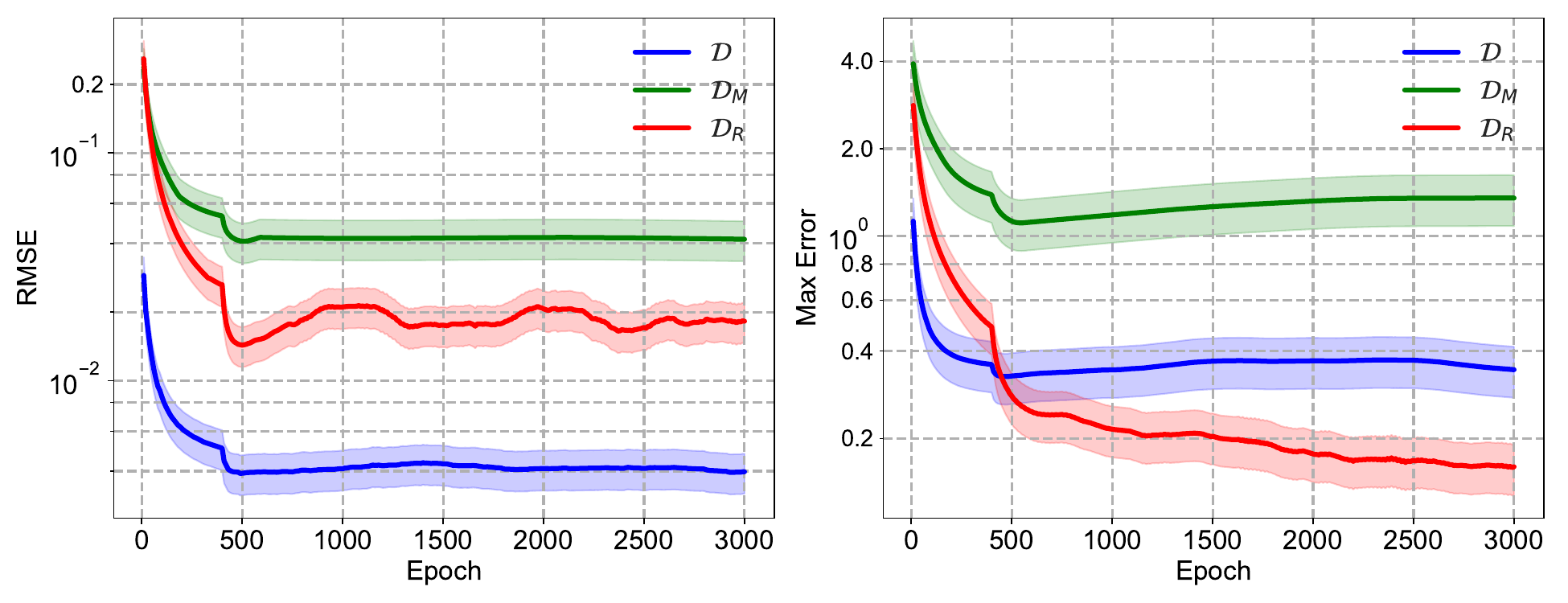}
      \caption{Quadcopter}
      \label{fig:quadcopter_error}
  \end{subfigure}
  \caption{Error metrics for MLP models trained on datasets from physical systems.}
  \label{fig:real_system_error}
  \vskip -0.1in
\end{figure}

Finally, Figures~\ref{fig:cartpole_error} and~\ref{fig:quadcopter_error} provide a detailed analysis of the prediction errors for multi-layer perceptron (MLP) models trained on the $\mathcal{D}$, $\mathcal{D}_M$, and $\mathcal{D}_R$ datasets from the Cartpole and Quadcopter environments. The $\mathcal{D}_R$ dataset, processed with EDS, shows lower maximum prediction errors despite slightly higher root-mean-square error (RMSE), indicating improved robustness under adverse conditions. This highlights the effectiveness of EDS in enhancing model performance and reliability in the presence of noisy and imbalanced data.

In summary, our evaluation shows that EDS enhances data coverage and model robustness, effectively addressing noise and imbalance to improve performance and generalization in real-world environments.

\section{Discussion and Conclusion}

In this paper, we introduced Error Distribution Smoothing (EDS) to address low-dimensional imbalanced regression by adjusting sample density based on the local Complexity-to-Density Ratio (CDR), ensuring balanced representation across the feature space. Experiments on simulated and real-world physics datasets demonstrated that EDS significantly improves precision, efficiency, and robustness. By reducing redundant samples in low-CDR regions and preserving critical points in high-CDR areas, EDS notably reduced the maximum prediction error, shortened training times through an optimized dataset size, and maintained model robustness despite varying data complexity. These results underscore EDS's effectiveness in overcoming the limitations of traditional sampling methods, enhancing data representation and model performance for imbalanced datasets.

\textbf{Future work.} We will focus on extending EDS to higher-dimensional datasets while optimizing its algorithms to significantly improve processing speed for large and complex datasets. This involves developing methods to handle increased complexity efficiently and exploring techniques such as parallel processing to enhance computational performance, ensuring EDS remains effective and scalable for real-world applications.

\textbf{Impact Statements.} This paper presents work whose goal is to advance the field of Machine Learning. There are many potential societal consequences of our work, none which we feel must be specifically highlighted here.

\bibliography{references}

\newpage
\appendix
\onecolumn

\section{Analysis of Linear Interpolation}
\label{appendix:LIMs}

\subsection{MISO Function Approximation via Linear Interpolation}

In this subsection, we explore the approximation of Multi-Input Single-Output (MISO) functions using linear interpolation over $n$-dimensional simplices. A MISO function maps multiple input variables to a single output variable, and linear interpolation provides a computationally efficient method for approximating such functions within a discretized feature space. By leveraging the geometric properties of simplices, we can construct accurate local models that facilitate efficient and robust function approximation.

Let us consider an $n$-dimensional simplex $\Omega_q$ with vertices $\{\boldsymbol{x}_{q,0}, \boldsymbol{x}_{q,1}, \ldots, \boldsymbol{x}_{q,n}\}$. For any point $\boldsymbol{x} \in \Omega_q$, its barycentric coordinates $\lambda_{q,i}(\boldsymbol{x})$ are uniquely defined as the solution to the following system:

\begin{equation}
\begin{cases}
  \sum_{i=0}^{n} \lambda_{q,i}(\boldsymbol{x}) = 1 \\
  \sum_{i=0}^{n} \lambda_{q,i}(\boldsymbol{x}) \boldsymbol{x}_{q,i} = \boldsymbol{x}
\end{cases}
\end{equation}

These coordinates satisfy the Kronecker delta property: $\lambda_{q,i}(\boldsymbol{x}_{q,j}) = \delta_{ij}$, which means that the coordinate corresponding to a vertex is one when evaluated at that vertex and zero elsewhere.

The barycentric coordinate $\lambda_{q,i}(\boldsymbol{x})$ for any point $\boldsymbol{x}$ relative to vertex $\boldsymbol{x}_{q,i}$ can be calculated without singling out any particular vertex by:

\begin{equation}
\lambda_{q,i}(\boldsymbol{x}) = \frac{\det\left(
\begin{bmatrix}
  \boldsymbol{x}_{q,1} - \boldsymbol{x}_{q,0} & \cdots & \boldsymbol{x}_{q,n} - \boldsymbol{x}_{q,0} & \boldsymbol{x} - \boldsymbol{x}_{q,0}
\end{bmatrix}^{\hat{i}}
\right)}{\det\left(
\begin{bmatrix}
  \boldsymbol{x}_{q,1} - \boldsymbol{x}_{q,0} & \cdots & \boldsymbol{x}_{q,n} - \boldsymbol{x}_{q,0}
\end{bmatrix}
\right)}
\end{equation}

where $\left[\cdot\right]^{\hat{i}}$ denotes the matrix formed by omitting the $i$th column from the original matrix.To estimate the error in linear interpolation, we use the second-order Taylor expansion of $f(\boldsymbol{x})$ around a point $\boldsymbol{x}$ within $\Omega_q$:

\begin{equation}
f(\boldsymbol{x}_{q,i}) = f(\boldsymbol{x}) + \nabla f(\boldsymbol{x})^T (\boldsymbol{x}_{q,i} - \boldsymbol{x}) + \frac{1}{2} (\boldsymbol{x}_{q,i} - \boldsymbol{x})^T \boldsymbol{H}(\xi_i) (\boldsymbol{x}_{q,i} - \boldsymbol{x})
\end{equation}

Given the function $f$ and points $\boldsymbol{x}_{q,i}$ with associated weights $\lambda_{q,i}$, under the normalization condition $\sum_{i=0}^{n} \lambda_{q,i} = 1$ and the centroid condition $\sum_{i=0}^{n} \lambda_{q,i} \boldsymbol{x}_{q,i} = \boldsymbol{x}$ \cite{berrut_barycentric_2004}, we examine the expression:

\begin{equation}
\sum_{i=0}^{n} \lambda_{q,i} f(\boldsymbol{x}_{q,i}) - f(\boldsymbol{x})
\end{equation}

Expanding this using the Taylor series gives:

\begin{equation}
\sum_{i=0}^{n}\left[ \lambda_{q,i} f(\boldsymbol{x}) +  \nabla f(\boldsymbol{x})^T \lambda_{q,i} (\boldsymbol{x}_{q,i} - \boldsymbol{x}) + \frac{1}{2} (\boldsymbol{x}_{q,i} - \boldsymbol{x})^T \boldsymbol{H}(\xi_i) (\boldsymbol{x}_{q,i} - \boldsymbol{x}) \right] - f(\boldsymbol{x}),
\end{equation}

where the first-order term vanishes due to the centroid condition:

\begin{equation}
\nabla f(\boldsymbol{x})^T \sum_{i=0}^{n} \lambda_{q,i} (\boldsymbol{x}_{q,i} - \boldsymbol{x}) = \nabla f(\boldsymbol{x})^T (\boldsymbol{x} - \boldsymbol{x}) = 0
\end{equation}

Therefore, the error in linear interpolation is bounded by the second-order term, which cannot be proven to be zero using only the given conditions. The maximum interpolation error over the simplex $\Omega_q$  $x \in \Omega_q$can be bounded as follows:

\begin{equation}
  \begin{aligned}
  \left|\sum_{i=0}^{n} \lambda_{q,i} f(\boldsymbol{x}{q,i}) - f(\boldsymbol{x})\right| &= \left|\left|\sum{i=0}^{n} \left[ \frac{1}{2} \lambda{q,i} (\boldsymbol{x}{q,i} - \boldsymbol{x})^T \boldsymbol{H}(\boldsymbol{\xi}_i) (\boldsymbol{x}{q,i} - \boldsymbol{x}) \right]\right|\right|_2 \\
  &\leq \frac{1}{2} \sum_{i=0}^{n} \left[ \lambda_{q,i}\ |\boldsymbol{x}_{q,i} - \boldsymbol{x}\|_2^2 \cdot \|\boldsymbol{H}(\boldsymbol{\xi}_i)\|2 \right] \\
  &< \frac{1}{2} \max_{\boldsymbol{x} \in \Omega_q} \|\boldsymbol{H}(\boldsymbol{x})\|_2 \cdot \max_{\boldsymbol{x}_1, \boldsymbol{x}_2 \in \Omega_q} \|\boldsymbol{x}_1 - \boldsymbol{x}_2\|_2^2 \cdot \sum_{i=0}^{n} \lambda_{q,i} \\
  &= \frac{1}{2} \max_{\boldsymbol{x} \in \Omega_q} \|\boldsymbol{H}(\boldsymbol{x})\|_2 \cdot \max_{\boldsymbol{x}_1, \boldsymbol{x}_2 \in \Omega_q} \|\boldsymbol{x}_1 - \boldsymbol{x}_2\|_2^2 \\
  &\leq \frac{1}{2} \max_{\boldsymbol{x} \in \Omega_q} \|\boldsymbol{H}(\boldsymbol{x})\|_F \cdot \max_{\boldsymbol{x}_1, \boldsymbol{x}_2 \in \Omega_q} \|\boldsymbol{x}_1 - \boldsymbol{x}_2\|_2^2
  \end{aligned}
\end{equation}
where $\|\boldsymbol{H(x)}\|_F$ is the Frobenius norm of the Hessian matrix of $f$ evaluated over $\Omega_q$, and $\|\boldsymbol{x} - \boldsymbol{y}\|_2$ is the Euclidean distance between points $\boldsymbol{x}$ and $\boldsymbol{y}$ within the simplex. This inequality provides an upper limit on the interpolation error, indicating that the error depends on the maximum curvature of $f$ over $\Omega_q$ and the size of the simplex.

The interpolation error for a function $f$ approximated via linear interpolation over an $n$-dimensional simplex $\Omega_q$ can be bounded above by:
\begin{equation}
\ub_{\boldsymbol{x} \in \Omega_q} | f(\boldsymbol{x}) - I_q(\boldsymbol{x})| 
= \frac{1}{2} \max_{\boldsymbol{x} \in \Omega_q} \|\boldsymbol{H}(\boldsymbol{x})\|_F \cdot \max_{\boldsymbol{x}_1, \boldsymbol{x}_2 \in \Omega_q} \|\boldsymbol{x}_1 - \boldsymbol{x}_2\|_2^2
\end{equation}
where $I_q(\boldsymbol{x})$ denotes the linear interpolated value at point $\boldsymbol{x}$, obtained as a weighted sum of function values at the simplex vertices with weights given by the barycentric coordinates. This inequality provides an upper limit on the interpolation error, highlighting its dependence on both the function's curvature and the simplex's spatial extent.

\subsection{MIMO Function Approximation via Linear Interpolation}

When considering the approximation of a Multiple Input Multiple Output (MIMO) function $\boldsymbol{f}(\boldsymbol{x}) = [f_1(\boldsymbol{x}), \ldots, f_m(\boldsymbol{x})]^\top$, where $\boldsymbol{x}$ is the feature vector and $m$ denotes the dimension of the output vector, each component function $f_i(\boldsymbol{x})$ can be treated as a MISO (Multiple Input Single Output) function with its own Hessian matrix $\boldsymbol{H}_i(\boldsymbol{x})$. The collection of these Hessian matrices forms a Hessian tensor for the MIMO function $\boldsymbol{f}(\boldsymbol{x})$.

The interpolation error for each individual MISO function $f_i(\boldsymbol{x})$ over a space $\Omega_q$ in the feature space can be bounded. Therefore, for each $f_i(\boldsymbol{x})$, the maximum interpolation error can be bounded as:

\begin{equation}
\ub_{\boldsymbol{x} \in \Omega_q} | f_i(\boldsymbol{x}) - I_{q,i}(\boldsymbol{x})| 
\leq \frac{1}{2} \max_{\boldsymbol{x} \in \Omega_q} \|\boldsymbol{H}_i(\boldsymbol{x})\|_F \cdot \max_{\boldsymbol{x}_1, \boldsymbol{x}_2 \in \Omega_q} \|\boldsymbol{x}_1 - \boldsymbol{x}_2\|_2^2
\end{equation}

where $I_{q,i}(\boldsymbol{x})$ denotes the interpolated value of $f_i(\boldsymbol{x})$ at point $\boldsymbol{x}$, $\|\boldsymbol{H}_i(\boldsymbol{x})\|_F$ is the Frobenius norm of the Hessian matrix of $f_i$ at $\boldsymbol{x}$, and $\|\boldsymbol{x}_1 - \boldsymbol{x}_2\|_2$ represents the Euclidean distance between points $\boldsymbol{x}_1$ and $\boldsymbol{x}_2$ within the simplex.

To find an upper bound on the approximation error for the entire MIMO function $\boldsymbol{f}(\boldsymbol{x})$, we consider the Euclidean norm of the error vector. For a MIMO function, the Hessian tensor's contribution to the error must be aggregated across all outputs. However, because it is not straightforward to define a single norm for a higher-order tensor, we typically resort to bounding the error based on the norms of the individual Hessian matrices:

\begin{equation}
\sqrt{\sum_{i=1}^{m}|f_i(\boldsymbol{x}) - I_{q,i}(\boldsymbol{x})|^2} \\
\leq \sqrt{\sum_{i=1}^{m}\left(\frac{1}{2} \max_{\boldsymbol{x} \in \Omega_q} \|\boldsymbol{H}_i(\boldsymbol{x})\|_F \cdot \max_{\boldsymbol{x}_1, \boldsymbol{x}_2 \in \Omega_q} \|\boldsymbol{x}_1 - \boldsymbol{x}_2\|_2^2\right)^2}
\end{equation}

By applying the Cauchy-Schwarz inequality to the non-negative terms, we derive a tighter upper bound on the approximation error:

\begin{equation}
  \begin{aligned}
    \|\boldsymbol{f(x)} - I_q(\boldsymbol{x})\|_2 &= \sqrt{\sum_{i=1}^{m}|f_i(\boldsymbol{x}) - I_{q,i}(\boldsymbol{x})|^2} \\
    &\leq \frac{1}{2} \max_{\boldsymbol{x}_1, \boldsymbol{x}_2 \in \Omega_q} \|\boldsymbol{x}_1 - \boldsymbol{x}_2\|_2^2 \cdot \sqrt{\sum_{i=1}^{m}\left(\max_{\boldsymbol{x} \in \Omega_q} \|\boldsymbol{H}_i(\boldsymbol{x})\|_F\right)^2} \\
    &\leq \frac{1}{2} \max_{\boldsymbol{x}_1, \boldsymbol{x}_2 \in \Omega_q} \|\boldsymbol{x}_1 - \boldsymbol{x}_2\|_2^2 \cdot \max_{\boldsymbol{x} \in \Omega_q} \|\boldsymbol{H}(\boldsymbol{x})\|_F
  \end{aligned}
\end{equation}

This expression sets an upper limit for the interpolation error of the MIMO function $\boldsymbol{f}(\boldsymbol{x})$ within simplex $\Omega_q$. It shows how the error depends on three factors: the number of outputs $m$, the maximum curvature of each component function in $\Omega_q$, and the size of the simplex.

\section{Error Distribution Smoothing Convergence Analysis}

\subsection{Assumptions and Notations}
We begin by formalizing the assumptions that underpin our analysis, using a rigorous mathematical notation typical of machine learning literature. We denote $\mathcal{D} = \{(\boldsymbol{x}_i, \boldsymbol{y}_i)\}_{i=1}^N$ as the dataset, where each pair $(\boldsymbol{x}_i, \boldsymbol{y}_i)$ is drawn from an unknown probability distribution $P(x)$ over the feature space $\mathcal{X}$.

The following assumptions are made regarding the data distribution, smoothness of the target function, sample size sufficiency, and uniformity within simplexes:

\begin{itemize}
    \item \textbf{Data Distribution Assumption:} The dataset $\mathcal{D}$ is assumed to be sampled i.i.d. from a distribution with a strictly positive density function over $\mathcal{X}$:
    \begin{equation}
    P(\boldsymbol{x}) > 0, \quad \forall \boldsymbol{x} \in \mathcal{X}.
    \end{equation}
    
    \item \textbf{Smoothness Assumption:} The target function $f: \mathcal{X} \to \mathbb{R}$ is assumed to be twice continuously differentiable, with its Hessian matrix having a bounded Frobenius norm:
    \begin{equation}
    \|\boldsymbol{H(x)}\|_F \leq M, \quad \forall \boldsymbol{x} \in \mathcal{X},
    \end{equation}
    where $M \in \mathbb{R}^{+}$ is a constant.
    
    \item \textbf{Sufficient Sample Size Assumption:} As the dataset size $N \to \infty$, the empirical distribution converges in probability to the true distribution:
    \begin{equation}
    \lim_{N \to \infty} \frac{1}{N} \sum_{i=1}^{N} \delta_{\boldsymbol{x}_i}(\Omega) = \int_\Omega  P(\boldsymbol{x}) dx, \quad \forall \Omega \subseteq \mathcal{X},
    \end{equation}
    with $\delta_{\boldsymbol{x}_i}$ being the Dirac delta function centered at $\boldsymbol{x}_i$.
    
    \item \textbf{Uniformity within Simplexes Assumption:} Within sufficiently small simplexes $\Omega$, data points are considered locally uniformly distributed. For any measurable subset $B \subseteq \Omega$, the following holds:
    \begin{equation}
    P(B) = \frac{\lambda(B)}{\lambda(\Omega)},
    \end{equation}
    where $\lambda(\cdot)$ represents the Lebesgue measure.
\end{itemize}

\subsection{Expectation Over Simplices and Size Function Properties}
When a point is added within a simplex $\Omega$, it divides into $n+1$ new simplices. By the Radon-Nikodym theorem, we have:
\begin{equation}
\mathbb{E}[\lambda_i] = \frac{\lambda(\Omega)}{n+1}, \quad i = 1, 2, \dots, n+1
\end{equation}

A size function $g_s(\Omega)$ is defined based on geometric properties and has a statistical relationship with $\lambda(\Omega)$, quantifying the maximum squared Euclidean distance between any two points in $\Omega$:
\begin{equation}
g_s(\Omega) = \max_{\boldsymbol{x}_1, \boldsymbol{x}_2 \in \Omega} \|\boldsymbol{x}_1 - \boldsymbol{x}_2\|_2^2
\end{equation}
The expected value of $g_s(\Omega_i)$ scales with the Lebesgue measure of $\Omega_i$ and $\Omega$:
\begin{equation}
\mathbb{E}[g_s(\Omega_i)] \propto \mathbb{E}[\lambda(\Omega_i)]^{\frac{2}{n}} = \mathbb{E}[\lambda(\Omega)]^{\frac{2}{n}} \cdot (n + 1)^{-\frac{2}{n}}
\end{equation}

\subsection{Convergence of Error Bounds}

In n-dimensional Euclidean space, each insertion divides simplices into $n+1$ smaller ones, reducing their volume by a factor of $(n+1)^{-1}$. Assume that at each iteration $k+1$, we insert $|\mathcal{F}_k|$ data pairs into the representative dataset, where $\mathcal{F}_k$ denotes the set of non-overlapping regions partitioned from the feature space at the $k$-th iteration. The error decreases exponentially with each iteration, specifically:
\begin{equation}
\mathbb{E}[\psi_{k+1}] = \mathbb{E}[\psi_{k}] \cdot (n+1)^{-\frac{2}{n}}.
\end{equation}

For low dimensions ($n$ small), the term $(n+1)^{-\frac{2}{n}}$ is significantly less than 1, leading to rapid convergence towards zero. Therefore, in low-dimensional settings, the error bounds decrease rapidly:
\begin{equation}
\lim_{k \to \infty} \mathbb{E}[\psi_k] = \lim_{k \to \infty} \mathbb{E}[\psi_0] (n+1)^{-\frac{2k}{n}} = 0
\end{equation}

In high-dimensional scenarios, while the limit still holds, the rate of convergence slows down because $(n+1)^{-\frac{2}{n}}$ approaches 1 as $n \to \infty$, indicating slower reductions in the upper bound of prediction errors.

To quantify this effect, we consider the limit:
\begin{equation}
\begin{aligned}
  \lim_{n \to \infty} (n+1)^{-\frac{2}{n}} &= \exp \left\{\lim_{n \to \infty} -2 \frac{\ln(n+1)}{n}\right\} \\
  &= \exp \left\{\lim_{n \to \infty} -2 \frac{\ln(n+2) - \ln(n+1)}{n+1-n}\right\} \text{(O.Stolz theorem)} \\
  &= \exp \left\{\lim_{n \to \infty} -2 \ln\left(1+\frac{1}{n+1}\right)\right\} \\
  &= \exp \left\{0\right\} \\
  &= 1
\end{aligned}
\end{equation}

The impact of dimensionality on the convergence rate is illustrated by the formula $(n+1)^{-\frac{2}{n}}$, which quantifies how the convergence rate changes with increasing dimensions $n$. For instance, when $n=2$, the value is approximately $0.3333$; for $n=5$, it increases to about $0.4886$; at $n=10$, it further rises to around $0.6190$; and for higher dimensions such as $n=50$ and $n=100$, the values approach $0.8545$ and $0.9118$, respectively. These results establish the theoretical foundation for the convergence of error bounds and provide a rigorous connection between the measure $\lambda(\Omega)$ and the size function $g_s(\Omega)$. This ensures that our analysis remains both robust and theoretically sound.

\section{Experimental Settings}
All of the following datasets were standardized. Specifically, for the full dataset, the mean $\mu_f$ and variance $\sigma_f$ of the features, as well as the mean $\mu_l$ and variance $\sigma_l$ of the labels, are computed. For each sample, the features $X_i$ and labels $Y_i$ are standardized using the following formulas:
\begin{equation}
  X_i' = \frac{X_i - \mu_f}{\sigma_f}, \quad Y_i' = \frac{Y_i - \mu_l}{\sigma_l}
\end{equation}
Regression tasks are performed on this standardized dataset.

\subsection{Motivation Example}
\textbf{Dataset Description} 

The generated dataset is designed for illustrating the principles of Error Distribution Smoothing(EDS). The dataset contains two features ($\boldsymbol{x_1, x_2}$) and a single label ($\boldsymbol{y}$), constructed using the following function:
\begin{equation}
  y = \frac{1}{0.33 + x_1^2 + x_2^2}
\end{equation}
where the features $x_1$ and $x_2$ are independently sampled from a uniform distribution within the range $\textbf{[-3, 3]}$. The dataset is divided into two subsets: a training set consisting of \textbf{5,000} samples and a test set with \textbf{5,000} samples. To ensure reproducibility, a fixed random seed is used during data generation.

\textbf{Regressor Description} 

We employ the Multilayer Perceptron (MLP) model for the regression task. The model consists of \textbf{3} hidden layers, each with \textbf{64} neurons and the \texttt{SiLU} activation function. The input layer accepts two features, and the output layer produces a single value. The network is implemented using PyTorch and trained using the \textbf{LBFGS} optimizer with a learning rate of 1 and a maximum iteration of 30 per step. The termination tolerance on first order optimality of LBFGS is $10^{-7}$ and the termination tolerance on function value changes is $10^{-9}$. We utilize the \textbf{Mean Squared Error (MSE) loss} throughout the experiments and fix the batch size to the size of the entire dataset for each epoch.The model is trained for 200 epochs.

\begin{table}[h!]
\centering
\renewcommand{\arraystretch}{1.5} 
\caption{Motivation Example Regression Configuration}
\begin{tabular}{cccccccc}
\hline
Model & Hidden Layer & Hidden Size & Activation Function & Optimizer & Learning Rate & Evaluation & Epochs\\
\hline
MLP & 3 & 64 & \texttt{SiLU} & L-BFGS & 1 & MSE & 200\\
\hline
\end{tabular}
\end{table}

%

\subsection{Dynamics System Identification}
\textbf{Dataset Description} 

The dataset is designed to evaluate EDS in dynamics system identification and we choose a discretized \textbf{Lorenz} system as an example. It includes three independent features ($ \textbf{x, y, z} $) and three labels ($ \boldsymbol{\dot{x},\dot{y},\dot{z}}$), calculated using the following system of equations:
\[
\begin{aligned}
    \dot{x} &= \sigma (y - x), \\
    \dot{y} &= x (\rho - z) - y, \\
    \dot{z} &= x y - \beta z,
\end{aligned}
\]
where $\sigma = 10$, $\rho = 28$, and $\beta = -\frac{8}{3}$. 

The dataset is generated by numerically solving the equations using the fourth-order Runge-Kutta method (RK4), where the initial $(x_0,y_0,z_0)$ are uniformly generated within the range $\textbf{[-10,10]}$ with 30 points, and the RK4 time step is \textbf{0.02} with a maximum time length of \textbf{20}. Both the train set and the test set consists of \textbf{15,000} samples. To ensure reproducibility, a fixed random seed is used during data generation.

\textbf{Regressor Description}

We employ the \textbf{Sparse Identification of Nonlinear Dynamics (SINDy)} model for the regression task. The model identifies the governing equations of the system by combining a library of candidate functions with Lasso regression to enforce sparsity in the identified coefficients.

The feature library is constructed using polynomial combinations of the input features ($x, y, z$). The library includes terms up to a \textbf{polynomial order} of 2, resulting in terms such as $1$, $x$, $y$, $z$, $x^2$, $xy$, and so on. Lasso regression is employed to determine the coefficients. The regularization parameter $\boldsymbol{\alpha=0.01}$ controls the sparsity of the solution and the polynomial order is set to 2. Also, the maximum iterations of Lasso regression is 10,000 with a tolerance of $10^{-6}$ for convergence.

The model is trained separately for each output dimension ($\dot{x}, \dot{y}, \dot{z}$) using the training set which the training data contains \textbf{300} samples and the test data contains \textbf{1,000} samples. The model's performance is evaluated using \textbf{Root Mean Squared Error (RMSE)} and \textbf{Maximum Error (Max Error)}.

\begin{table}[h!]
\centering
\renewcommand{\arraystretch}{1.5} 
\caption{Dynamics System Identification Regression Configuration}
\begin{tabular}{cccccc}
\hline
Regression & Regulatization Para & Tolerance & Evaluation\\
\hline
SINDy & $\alpha = 0.01$ & $10^{-6}$ & RMSE, Max Error\\
\hline
\end{tabular}
\end{table}

\subsection{Polar Moment of Inertia}
\textbf{Dataset Description} 

The dataset is generated to evaluate EDS in high-dimensional settings. The dataset contains \textbf{50,000} samples, each representing a rectangle with a white outline on a black background. The images are grayscale with dimensions $280 \times 280$, normalized to the range $[0, 1]$, where white pixels are represented as $L(i, j) = 1$ and black pixels as $L(i, j) = 0$. 

For each rectangle, two types of features are extracted. First, \textbf{high-dimensional} features are derived from the flattened pixel values of the grayscale image. Second, \textbf{low-dimensional} features consist of the coordinates of the rectangle's bottom-left corner $(x_1, y_1)$ and top-right corner $(x_2, y_2)$. 

The labels are calculated as the polar moment of inertia $\boldsymbol{J}$ relative to the centroid, using the formula
\begin{equation}
  J = \sum_{i=1}^{W} \sum_{j=1}^{H} (i^2 + j^2) L(i, j)
\end{equation}
where $W$ and $H$ denote the image width and height. The computed values are scaled by a factor of $10^8$ to facilitate numerical stability.

The dataset is split into a training set containing \textbf{10,000} samples and a testing set containing \textbf{40,000} samples. Both high-dimensional features (pixel values) and low-dimensional features (coordinates) are saved in text files, along with their corresponding labels. The regression task is to predict the polar moment of inertia based on the geometric and pixel-based features of the rectangles.

\textbf{Regressor Description}

A fully connected Multilayer Perceptron (MLP) model is used for regression, featuring 2 hidden layers with 128 neurons each and a \texttt{Tanh} activation function. The input consists of four dimensions: the coordinates $(x_1, y_1)$ and $(x_2, y_2)$, with a single output dimension representing the polar moment of inertia $J$. Training employs the \texttt{Adam} optimizer with a learning rate of 0.0005 and regularization parameter $\alpha = 0.0001$, using a batch size of 128.

The training process includes a convergence tolerance of $10^{-8}$, momentum set to 0.9 with Nesterov's acceleration, and a fixed random seed of 3704 for reproducibility. Validation is performed on 10\% of the data, and training stops if there's no improvement in validation score over 10 consecutive iterations, or after 15,000 function evaluations. Performance is assessed using RMSE and Maximum Error metrics on standardized data.

\begin{table}[htbp]
  \centering
  \renewcommand{\arraystretch}{1.5} 
  \caption{Polar Moment of Inertia Regression Configuration}
  \begin{tabular}{ccccccccc}
  \hline
  Model & Hidden Layer & Hidden Size & \makecell{Activation\\Function} & Optimizer & Regularization & \makecell{Learning\\Rate} & Evaluation \\
  \hline
  MLP & 2 & 64 & \texttt{Tanh} & \texttt{Adam} & $\alpha = 10^{-4}$ & 0.0005 & \makecell{RMSE,\\ Max Error} \\
  \hline
  \end{tabular}
\end{table}
  
\subsection{Real-World Experiment}
\subsubsection{Cartpole}
\textbf{Dataset Description}

The Cartpole is specifically the single linear IP570 model, designed for STM32 microcontrollers. The dimensions of the system are $560\text{mm} \times 136\text{mm} \times 328\text{mm}$ and the total weight is 2.4kg (including the pendulum). The main control chip is the STM32F103C8T6, and the motor used is the MG513. The travel distance of the cart is 395mm, and the system includes two fixed bases. We utilized a \textbf{PID} controller to control the cartpole.

For the dataset, the features are angle ($\theta$) and angular velocity ($\omega$) of the pendulum rod, as well as the control output from the controller. The labels are the acceleration of the cart and the angular acceleration of the pendulum ($\beta$). There are \textbf{13,972} samples in the dataset of which 75\%(\textbf{10,479}) is used as training set and 25\%(\textbf{3,493}) as test set.

\textbf{Regressor Description}

We also employ the MLP model for the regression task. The input layer accepts 3 features and the output layer produces 2 labels. Other configurations are the same as in experiment Polar Moment of Inertia

\subsubsection{Quadcopter}
\textbf{Dataset Description}

We collected data using a centrally symmetric quadcopter with a takeoff weight of $1.40\,\text{kg}$, rotor radius of $12.0\,\text{cm}$, and wheelbase of $35.0\,\text{cm}$. The flight control system is powered by Pixhwak4 (PX4), which outputs PWM signals to adjust the effective voltage of the motors and thus regulate the motor speeds. 

Height $(h)$, throttle commands ($T$, related to the motor speed), and velocity $v$ are selected as features, while the longitudinal acceleration $ (a) $ of the quadcopter serves as the label. The dataset is processed using Kalman smoother and standardized. \textbf{100} flight experiments are conducted, with data collection at $\boldsymbol{0.05}\,\text{s}$ intervals for $\boldsymbol{30}\,\text{s}$. The deisred altitude is set as:
\begin{equation}
  h_{d} = h_{0} + A \sin(2\pi f t)
\end{equation}
where $h_{0} = 3.0\,\text{m}$. For each experiment, $\boldsymbol{A}$ and $\boldsymbol{f}$ are uniformly sampled from [$\boldsymbol{1, 2}$] and [$\boldsymbol{0.005, 0.02}$], respectively. Both the train set and teh test set have \textbf{30,000} samples.

\textbf{Regressor Description}

We still use the MLP model for the regression task, The input layer accepts 3 features and the output layer produces a single value. And the other parameters are the same as before.

All the above experimental settings are summarized in the following table.
\begin{table}[htbp]
\centering
\renewcommand{\arraystretch}{1.5} 
\caption{Experimental Settings}
\begin{tabular}{lcccccc}
\hline
Dataset & Features & Label(s) & Train set & Test set  &  Regressor\\
\hline
Motivation Example & $x_1,x_2$ & $y$ & 5,000 & 1,000 & MLP\\
Lorenz System & $x,y,z$ & $\dot{x},\dot{y},\dot{z}$ & 15,000 & 15,000 & SINDy\\
Polar Moment of Inertia & $x_1, y_1, x_2, y_2$ & $J$ & 10,000 & 40,000 & MLP\\
Cartpole & $\theta,\omega, I$ & $a_{cart}, a_{pole}$ & 10,479 & 3,493 & MLP\\
Quadcopter & $h, v, T$ & $a_z$ & 30,000 & 30,000 & MLP\\
\hline
\end{tabular}
\end{table}

\end{document}